\documentclass[letterpaper, 10 pt, conference]{formatting/ieeeconf}
\IEEEoverridecommandlockouts                              
\let\labelindent\relax
\usepackage{amsfonts}       

\usepackage{amsthm}
\usepackage{wrapfig}
\usepackage{mathtools}      
\usepackage{amssymb}        
\usepackage{filecontents}
\usepackage{graphicx}       
\usepackage{marginnote}     
\usepackage{marvosym}       
\usepackage{overpic}        
\usepackage{tabularx}
\usepackage{cite}
\usepackage{color}
\usepackage[linesnumbered,algoruled,boxed,lined]{algorithm2e}
\usepackage[normalem]{ulem}
\usepackage{epstopdf}
\usepackage{float}
\usepackage{enumitem}
\usepackage[normalem]{ulem}
\usepackage{lipsum}
\usepackage[a-2b,mathxmp]{pdfx}[2018/12/22]

\newif\ifdraft
\draftfalse

\ifdraft
\usepackage[paperheight=11in,paperwidth=9.5in,
			left=1.25in,right=1.25in,
			top=0.75in,bottom=0.75in,
			heightrounded,marginparwidth=1.2in,
			marginparsep=0.05in]{geometry}
\usepackage{xcolor}
\usepackage{xargs} 
\usepackage[textsize=footnotesize]{todonotes}
\newcommandx{\nt}[2][1=]{\todo[linecolor=red,
			backgroundcolor=red!10,bordercolor=red,#1]{ #2}}
\newcommandx{\jy}[2][1=]{\todo[linecolor=green,
			backgroundcolor=green!10,bordercolor=green,#1]{JY: #2}}
\else
\newcommand{\nt}[1]{{}}
\newcommand{\jy}[1]{{}}
\fi

\newif\iftwocolumn
\twocolumntrue

\theoremstyle{definition}

\theoremstyle{remark}

\SetKwProg{Fn}{Function}{}{}
\SetKwComment{Comment}{$\triangleright$\ }{}

\makeatletter
\def\subsubsection{\@startsection{subsubsection}
                                 {3}
                                 {\z@ \hspace*{1mm}}
                                 {0ex plus 0.1ex minus 0.1ex}
                                 {0ex}
                                 {\normalfont\normalsize\itshape}}
\makeatother

\newcommand{\mpp}{\textsc{MRPP}\xspace}

\newcommand{\pibt}{\textsc{PIBT}\xspace}
\newcommand{\clpibt}{\textsc{PBCR}\xspace}
\newcommand{\clcbs}{\textsc{CL-CBS}\xspace}
\newcommand{\clecbs}{\textsc{ECCR}\xspace}

\font\titlefont=ptmb at 15.1pt
\title{\titlefont
Decentralized Lifelong Path Planning for Multiple Ackerman Car-Like Robots
}
\author{Teng Guo   \qquad Jingjin Yu
\thanks{G. Teng, and J. Yu are with the Department of 
Computer Science, Rutgers, the State University of New Jersey, Piscataway, NJ, USA. 
Emails: {\tt\small \{teng.guo, jingjin.yu\}@rutgers.edu}.
}
}

\begin{document}

\maketitle
\thispagestyle{empty}
\pagestyle{empty}


\begin{abstract}
Path planning for multiple non-holonomic robots in continuous domains constitutes a difficult robotics challenge with many applications. 
Despite significant recent progress on the topic, computationally efficient and high-quality solutions are lacking, especially in lifelong settings where robots must continuously take on new tasks. 
In this work, we make it possible to extend key ideas enabling state-of-the-art (SOTA) methods for multi-robot planning in discrete domains to the motion planning of multiple Ackerman (car-like) robots in lifelong settings, yielding high-performance centralized and decentralized planners. 
Our planners compute trajectories that allow the robots to reach precise $SE(2)$ goal poses. 
The effectiveness of our methods is thoroughly evaluated and confirmed using both simulation and real-world experiments.
\end{abstract}

\section{Introduction}\label{sec:intro}
%
%
%
%

The rapid development of robotics technology in recent years has made possible many revolutionary applications. One such area is multi-robot systems, where many large-scale systems have been successfully deployed, including, e.g., in warehouse automation for general order fulfillment \cite{wurman2008coordinating}, grocery order fulfillment \cite{mason2019developing}, and parcel sorting \cite{wan2018lifelong}. However, upon a closer look at such systems, we readily observe that the robots in such systems largely live on some discretized grid structure. In other words, while we can effectively solve multi-robot coordination problems in grid-like settings, we do not yet see applications where many non-holonomic robots traverse smoothly in continuous domains due to a lack of good computational solutions. Whereas many factors contribute to this (e.g., state estimation), a major roadblock is the lack of efficient computational solutions tackling the lifelong motion planning for non-holonomic robots in continuous domains. 

Toward clearing the above-mentioned roadblock, in this work, we proposed two algorithms specially designed to solve static/one-shot and lifelong path/motion planning tasks for Ackerman car-like robots.
The basic idea behind our methods is straightforward: to enable the adaptation of discrete search strategies for car-like robots, we use a small set of fixed but representative motion primitives to transition the robots' states. These fixed motion primitives, when properly put together, yield near-optimal trajectories that ``almost'' connect the starts and goals for robots. Some local adjustments are then used to complete the full trajectory. 
The first algorithm and our main contribution in this research, \emph{Priority-inherited Backtracking for Car-like Robots} (\clpibt)  adapts a decentralized strategy that leverages search-prioritization strategies from Priority Inheritance and Backtracking (PIBT) \cite{okumura2019priority}. 
The second algorithm, \emph{Enhanced Conflict-based search for Car-like Robots} (\clecbs), is a centralized method building on the principles of Enhanced Conflict-Based Search (ECBS) \cite{barer2014suboptimal} and CL-CBS \cite{wen2022cl}, the car-like robot extension of (basic) conflict-based search \cite{sharon2015conflict}. 
We further boost algorithms' success rates by introducing carefully designed, effective heuristics which also reduce the occurrence of deadlocks.
Thorough simulation-based evaluations confirm that our methods deliver scalable SOTA performances on many key practical metrics. 
While our centralized methods tend to find shorter trajectories due to their access to global information, our decentralized method produces better scalability, yielding a higher success rate.

\begin{figure}[t]
    \centering
    \vspace{2.5mm}
  \begin{overpic}               
        [width=1\linewidth]{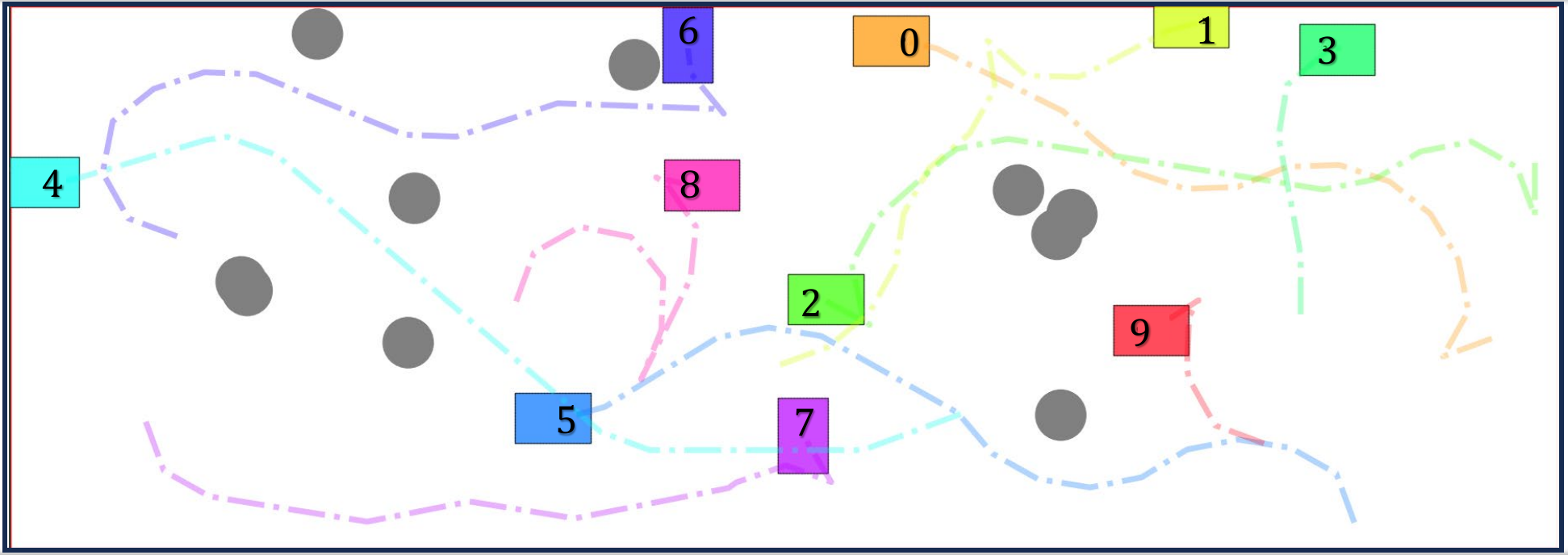}
             \small
        \end{overpic}
\vspace{-3mm}
    \caption{A simulated example on a $60\times 30$ map with 10 obstacles and robots. The collision-free trajectories for the car-like robots are shown. 
    }
    \label{fig:introduction}
\end{figure}%

\textbf{Related Work}. Multi-Robot Path Planning (\mpp) has garnered extensive research interests in robotics and artificial intelligence in general. The graph-based \mpp variant, also known as Multi-Agent Path Finding (MAPF) \cite{stern2019multi}, is to find collision-free paths for a set of robots within a given graph environment. Each robot possesses a distinct starting point, and a goal position, and the challenge lies in determining paths that ensure their traversal from start to goal without collisions.
It has been proven many times over that optimally solving the graph-based problem to minimize objectives like makespan or cumulative costs is NP-hard; see, e.g.,  \cite{surynek2010optimization,yu2013structure}. 

Computational methods for \mpp can be broadly classified into two categories: \emph{centralized} and \emph{decentralized}. Centralized solvers operate under the assumption that robot paths can be computed centrally and subsequently executed with minimal coordination errors. On the other hand, decentralized solvers capitalize on the autonomy of individual robots, enabling them to calculate paths independently while requiring coordination for effective decision-making.
Centralized solutions often involve reducing \mpp to well-established problems \cite{yu2016optimal,surynek2010optimization,erdem2013general}, employing search algorithms to explore the joint solution space \cite{sharon2015conflict,barer2014suboptimal,sharon2013increasing,silver2005cooperative,okumura2023lacam} or apply human-designed rules for coordination and collision-avoidance \cite{luna2011push, guo2022sub}. 
Decentralized approaches \cite{han2019ddm,okumura2019priority} leverage efficient heuristics to address conflicts locally, bolstering their success rates. Machine learning and reinforcement learning methods for \mpp have also started to emerge, with researchers putting forward data-driven strategies to directly learn decentralized policies for \mpp  \cite{sartoretti2019primal,li2020graph}.
Nevertheless, we note that the paths generated by graph-based \mpp algorithms cannot be directly applied to physical robots due to their disregard for robot kinematics. 

As autonomous driving gains momentum, interest in path planning for multiple car-like vehicles is also steadily rising \cite{Guo2023TowardEP,okoso2022high}. This has led to the development of new methods, including adaptations of CBS to car-like robots (CL-CBS) \cite{wen2022cl}, prioritized trajectory optimization \cite{li2020efficient}, sampling-based techniques \cite{le2019multi}, and optimal control strategies \cite{ouyang2022fast}. Decentralized approaches have been introduced as well, such as B-ORCA \cite{alonso2012reciprocal} and $\epsilon$CCA \cite{alonso2018cooperative}, both stemming from adaptations of ORCA \cite{blm-rvo} designed for car-like robots. These methods offer faster computation compared to centralized methods. However, their reliance on local information means they cannot provide a guarantee that robots will successfully reach their destinations. Moreover, they are prone to deadlock issues and tend to achieve much lower success rates, particularly in densely populated environments.

\textbf{Organization.}
The rest of the paper is organized as follows. 
Sec.~\ref{sec:problem} covers the preliminaries, including the problem formulation.
In Sec.~\ref{sec:pibt}-\ref{sec:ecbs}, we demonstrate our algorithms in detail.
In Sec.~\ref{sec:evaluation}, we conduct evaluations on the proposed methods in static settings and lifelong settings and discuss their implications.
We conclude in Sec.~\ref{sec:conclusion}.

%


%

%

\section{Problem Formulation}\label{sec:problem}
\subsection{Multi-Robot Path Planning for Car-Like Robots}
 A static instance of this problem is specified as $(\mathcal{W},\mathcal{S},\mathcal{G})$, where $\mathcal{W}$ constitutes a map with dimensions $W\times H$, housing a set of obstacles $\mathcal{O}=\{o_1,...,o_{n_o}\}$. $\mathcal{S}=\{s_1,...s_n\}$ defines the initial configurations of $n$ robots within the workspace, and $\mathcal{G}=\{g_1,...g_n\}$ represents the corresponding goal configurations.

Each robot's $SE(2)$ configuration, denoted as $v_i=(x_i,y_i,\theta_i)$, comprises a 3-tuple of position coordinates and the yaw orientation $\theta_i$. The car-like robot is modeled as a rectangular shape with length $\ell$ and width $w$. 
The subset of $\mathcal{W}$ occupied by a robot's body at a given state $v$ is denoted by $\Gamma(v)$. 
The motion of the robot adheres to the Ackermann-steering kinematics (see Fig.~\ref{fig:car_model}(a)):
\begin{equation}\label{eq:ackerman}
\dot{x}=u\cos\theta, \quad \dot{y}=u\sin\theta, \quad \dot{\theta}=\frac{u}{\ell_{b}}\tan\phi,
\end{equation}
in which $u\in[-u_{m},u_{m}]$ is the linear velocity of the robot, $\phi\in [-\phi_{m},\phi_{m}]$ is the steering angle. These are the control inputs. $\ell_b$ is the wheelbase length, i.e., the distance between the front and back wheels.

To make planning more tractable, we discretize into intervals $\Delta t$.
The goal is to ascertain a  feasible path for each robot $i$, expressed as a state sequence $P_i=\{p_i(0),...p_i(t),...p_i(T)\}$ that adheres to the following constraints:
(i) $p_i(0)=s_i$ and $p_i(T)=g_i$;
(ii) $\forall t\in [0,T], \forall i\neq j, \Gamma(p_i(t))\bigcap \Gamma(p_j(t))=\emptyset$;
(iii) The path follows the Ackermann-steering kinematic model.
The time interval is kept small during the discretization process. This choice ensures that the distance moved within each step remains smaller than the size of the robot. Consequently, the need to account for swap conflicts \cite{stern2019multi} is eliminated.

Trajectory quality is evaluated using the following criteria:
(i) Makespan: $\max_{1\leq i\leq n} len(P_i)/u_{max}$;
(ii) Flowtime: $\sum_{1\leq i\leq n}  len(P_i)/u_{max}$, where $len(P_i)$ denotes the length of trajectory $P_i$.

\subsection{The Lifelong Setting}
In a lifelong setting, we assume an infinite stream of tasks for each robot. Upon completing the current task, a robot immediately receives a new target.
In this scenario, evaluating the system's efficiency often relies on throughput—the number of tasks accomplished (or goal states arrived) within a specified number of timesteps.

\section{Priority-Inherited Backtracking for Car-Like Robots (\clpibt)}\label{sec:pibt}
In this section, we propose a \emph{decentralized} method called \clpibt for car-like robots, adapting the prioritization mechanism of the decentralized \mpp algorithm \pibt \cite{okumura2019priority}. 
In conjunction the introduction of \clpibt, we also describe the general methodology we adopt to plan continuous trajectories for non-holonomic robots that is generally applicable, provided that the optimal trajectories connecting the robot can be compactly represented using a few motion primitives. 

\subsection{Motion Primitives}
\begin{figure}[h]
    \centering
  \begin{overpic}               
        [width=1\linewidth]{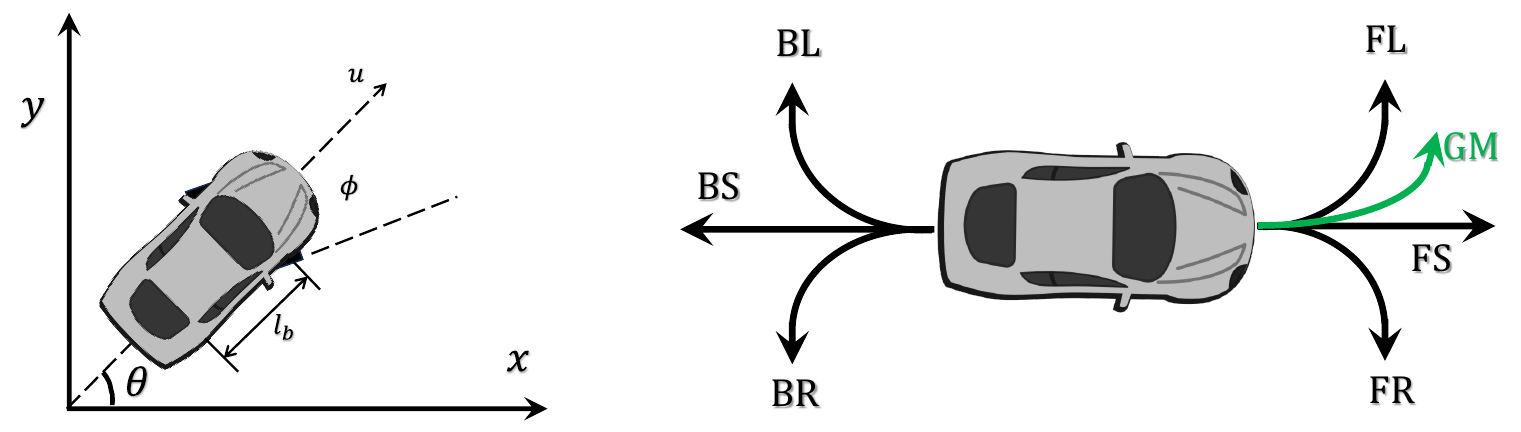}
             \small
             \put(18.5, -3) {(a)}
             \put(70.5, -3) {(b)}
        \end{overpic}
\vspace{-3mm}
    \caption{(a) Ackermann steering kinematic model. (b)  The predefined motion primitives. We have at most 8 motion primitives, which are
forward max-left (FL), forward straight (FS), forward max-
right (FR), backward max-left (BL), backward straight (BS),
backward max-right (BR),  wait, and greedy motion primitive (GM)}
    \label{fig:car_model}
\end{figure}

To render planning for the continuous system feasible, we constrain the robot's possible state transitions within the discretized time interval $\Delta t$, forming a set of \emph{motion primitives}.
Our motion primitives $\mathcal{M}$ are defined similarly as done in \cite{wen2022cl, li2020efficient}, which contain a total of seven actions: forward max-left (FL), forward straight (FS), forward
max-right (FR), backward max-left (BL), backward straight (BS), backward max-right (BR), and wait, as shown in Fig.~\ref{fig:car_model}(b).
For the first six motion primitives, we assume that the robot maintains a constant velocity of $u_m$ throughout a single time interval $\Delta t$. When the robot turns, we assume it is pivoting using the maximum steering angle $\phi_m$ and then tracing an arc with a turning radius $r_m$. This arc has a $u_{m}\Delta t$ length.
To ensure a robot can reach its goal state for \clpibt, one additional \emph{greedy motion primitive} (GM) is added to $\mathcal{M}$.
This greedy motion primitive is derived by truncating the first segment of length $u_{m}\Delta t$ from the shortest path connecting the current state to the goal state for a single robot. In the absence of obstacles on the map, this shortest path corresponds to the (optimal) Reeds-Shepp \cite{reeds1979optimal} path while if obstacles are present, the path can be determined using the vanilla (non-spatio-temporal) hybrid A* algorithm.
It's important to recognize that this greedy motion primitive could be identical to other motion primitives, resulting in the same subsequent state as those alternatives. In such cases, we opt to retain solely the greedy one.
\vspace{2mm}

\subsection{Algorithm Skeleton}
\clpibt is outlined in Alg.~\ref{alg:pibt_main}-\ref{alg:pibt_func},
at the core of which lies the \texttt{PIBTLoop}, which orchestrates the decision-making process for the robots at each time step $t$. \texttt{PIBTLoop}'s primary objective is to determine each robot's next motion and succeeding state without encountering collisions.

\texttt{PIBTLoop} initializes two essential sets: $\text{UNDECIDED}$ and $\text{OCCUPIED}$. 
The $\text{UNDECIDED}$ set keeps track of robots that have yet to finalize their next actions, while the $\text{OCCUPIED}$ set stores the information about occupied states, preventing multiple robots from attempting to occupy the same space simultaneously.
At the outset of each iteration, the algorithm updates the priorities of all robots based on relevant factors. 
The priorities are crucial in determining which robot takes precedence in decision-making. 
This step ensures that the higher-priority robots have their actions determined earlier, facilitating swift and efficient navigation.
\clpibt's decision-making process revolves around an iterative approach. The algorithm enters a loop that continues until all robots in the $\text{UNDECIDED}$ set have their actions determined.
The robot with the highest priority within the loop is selected from the $\text{UNDECIDED}$ set. This robot is designated as robot $i$ for the current iteration.
\begin{algorithm}
\begin{small}
\DontPrintSemicolon
\SetKwProg{Fn}{Function}{:}{}
\SetKw{Continue}{continue}
  \Fn{\textsc{PIBTLoop}()}{
 \caption{Priority selection at time step $t$ \label{alg:pibt_main}}
    $\text{UNDECIDED}\leftarrow \mathcal{R}(t)$\;
    $\text{OCCUPIED}\leftarrow \emptyset$\;
    update  all priorities\;
    \While{$\text{UNDECIDED}!=\emptyset$}{
        $i\leftarrow$ the robot with the highest priority in UNDECIDED\;
        $\texttt{PIBT}(i,\text{NONE},\text{NONE})$\;
    }
}
\end{small}
\end{algorithm}

The \texttt{PIBT} function is invoked for the selected robot $i$, along with placeholders for two parameters: the parent robot $j$ which robot $i$ is inherited from, and the potential succeeding state $v_j$ of the parent robot. This function serves as the primary interface for the robot's decision-making. It evaluates potential actions considering the robot's current state, goals, obstacles, and the presence of other robots. Based on these evaluations, the robot's next action is determined to ensure collision-free and goal-oriented navigation.

\begin{algorithm}
\begin{small}
\DontPrintSemicolon
\SetKwProg{Fn}{Function}{:}{}
\SetKw{Continue}{continue}
  \Fn{\textsc{PIBT}($i,j,v_j$)}{
 \caption{PIBT function  \label{alg:pibt_func}}
    $\text{UNDECIDED}\leftarrow \mathcal{R}(t)-i$\;
        $\text{NbrUndecided}\leftarrow\text{ the neighboring robots of } i\text{ that remain undecided}$\;
        $\text{NbrOcc}\leftarrow \text{the occupied states of the neighboring decided robots}$\;
    $C\leftarrow\{\texttt{ValidSuccState}(p_i(t), m_i)|m_i\in\mathcal{M}_i\}$\;

    \While{$C\neq \emptyset$}{
        $v_i\leftarrow \text{argmax}_{v\in C}Q_i(v)$ and remove $v_i$ from $C$\;
    
        \If{$\texttt{FindCollision}(v_i,\text{NbrOcc}\bigcup\{p_j(t),v_j\})$}{
            continue\;
        }
     $\text{OCCUPIED}.add(v_i)$\;
        \If{$\exists k\in \text{NbrUndecided}$ such that $\texttt{FindCollision}(p_k(t),v_i)\text{= true}$}{
        \If{$\texttt{PIBT}(k,i,v_i)$ is valid}{
            $p_i(t+1)\leftarrow v_i$\;
   
            \Return valid\;
        }
        \Else{    
            $\text{OCCUPIED}.remove(v_i)$\;
            }
        }
    }
   $\text{UNDECIDED}.add(i)$\;    
    \Return invalid\;
}
\end{small}
\end{algorithm}

The \texttt{PIBT} function plays a pivotal role within \clpibt, encapsulating the decision-making process for an individual robot. 
This function is called iteratively for each robot to compute its next action, considering its current state, goals, and interactions with neighboring robots.
The function identifies the set $\text{UNDECIDED}$, which comprises all robots yet to finalize their actions, and marks robot $i$ as decided temporarily. Additionally, $\text{NbrUndecided}$ contains neighboring robots of $i$ that remain undecided, and $\text{NbrOcc}$ holds the occupied states of the neighboring robots that have already determined their actions. These sets provide crucial contextual information to facilitate informed decision-making.
The function computes a set $C$ of potential future states $v_i$ for the current robot $i$. These states are generated based on all feasible actions $m_i$ available to robot $i$ at its current state $p_i(t)$. The goal is to explore various potential actions to lead to a successful next state for $r_i$.
The function selects the potential state $v_i$ within a loop that maximizes the robot's utility function $Q_i(v)$ from the set $C$. This action selection aims to optimize the robot's decision by choosing the most promising action according to the utility criterion.
However, before finalizing the action, the function checks for collisions between the selected $v_i$ and the occupied states of neighboring robots, as well as the state of the parent robot $j$ (if it is not NONE) and its potential succeeding state $v_j$. If a collision is detected, the function continues to the next iteration of the loop, considering alternative potential actions.

If no collisions are found for the chosen $v_i$, $v_i$ is added to the $\text{OCCUPIED}$ set, indicating the robot intends to occupy this state. The function then checks if an undecided neighboring robot $k$ will collide with $v_i$.
If such a robot $k$ is found, the function recursively calls \texttt{PIBT} for robot $k$ with the parent robot $i$ and the tentative state $v_i$ as the parameters. If the resulting action is valid, robot $i$ updates its next state $p_i(t+1)$ to $v_i$, and the function returns ``valid."
If \texttt{PIBT} for robot $k$ fails, the state $v_i$ is removed from $\text{OCCUPIED}$.
After evaluating all potential actions and considering interactions with neighboring robots, robot $i$ remains in the $\text{UNDECIDED}$ set. The function returns ``invalid" if a valid action cannot be determined. Otherwise, it returns ``valid" along with the updated next state $p_i(t+1)$ for robot $i$.

We note that \clpibt's one-step planning and execution nature makes it applicable to static and lifelong scenarios.

\subsection{Performance-Boosting Heuristics}
Multiple heuristics are introduced to boost the performance of \clpibt, while some are basic, others are involved. 

\textbf{Distance heuristic.} We use the max of the holonomic cost with obstacles, the length of the shortest Reeds-Shepp path between two states, and 2D Euclidean distance as our distance heuristic function (\texttt{DistH}) similar to \cite{dolgov2008practical,wen2022cl}.
This distance heuristic is admissible. 

\textbf{Priority heuristic.}
\clpibt constitutes a single-step, priority-driven planning approach that necessitates the ongoing adjustment of each robot's priority at each time step.
Initially, the priority assigned to each robot is determined based on the number of time steps that have transpired since its preceding task was updated.
The robot with more time since its previous goal update receives a higher priority ranking.
In static scenarios, the elapsed time is reset to zero whenever a robot reaches its designated goal state.
This reset mechanism prevents robots that have achieved their goal state from obstructing the progress of other robots.
When multiple robots share the same elapsed time, prioritization is resolved by favoring those robots with a larger heuristic distance value to their respective goal states.
This heuristic principle finds widespread application in the realm of prioritized planning \cite{van2005prioritized} to improve the success rate.

\begin{figure}[h]
    \centering
  \begin{overpic}               
        [width=1\linewidth]{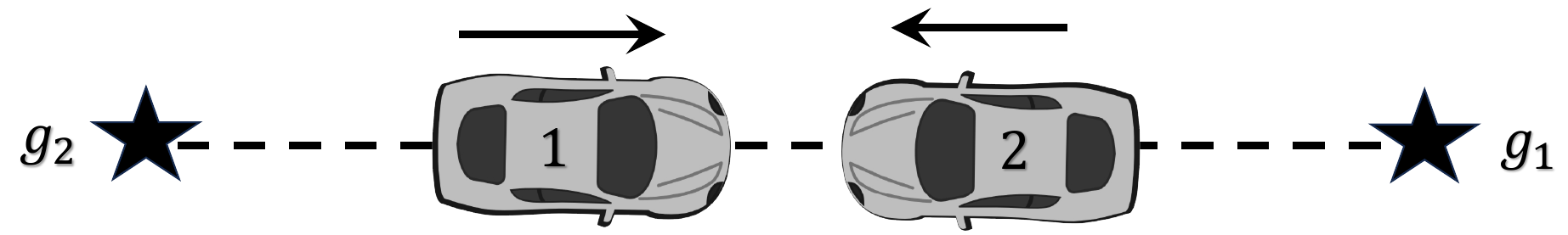}
        \end{overpic}
\vspace{-3mm}
    \caption{
An illustrative case highlighting the potential occurrence of deadlocks due to the exclusion of the count-based heuristic is as follows:
In this scenario, two robots, labeled as robot 1 and robot 2, travel in opposite directions along a linear path. When robot 2 yields to robot 1 in \clpibt, a predicament arises. In this situation, the motion primitives FS, FR, and FL lead to collisions, compelling robot 2 to consistently opt for the motion primitive BS. This choice is driven by BL and BR, involving additional turning penalties.
A dynamic shift occurs in priorities upon robot 1's successful arrival at its designated goal state. Robot 1 is assigned a lower priority than robot 2 and consequently yields to the latter. Similarly, robot 1 consistently selects BS using the greedy strategy. Consequently, an unending cycle emerges, entangling robot 1 and robot 2 in an indefinite sequence of movements }
    \label{fig:dead_lock}
\end{figure}

\textbf{The Q-function.}
The $Q$-function is a critical tool for assessing the optimal choice of a motion primitive in guiding the robot from its current state toward its goal state.
\pibt algorithm employs the single-agent shortest path length from the current vertex to its goal vertex, neglecting inter-agent collisions, as the basis for its evaluation function. This approach is feasible for discrete 2D graphs since these shortest path lengths can be pre-calculated and stored in a table. Nevertheless, this approach is impractical for the state space we're addressing with car-like robots. Firstly, the state space is infinite. Secondly, determining a single-robot shortest trajectory from a state to a goal state using hybrid A* for evaluating all the motion primitives is both time-intensive and incomplete.
For these reasons, we consider employing the distance heuristic function discussed earlier as well as the action cost, denoted as 
\vspace{1mm}
\begin{equation}\label{eq: Q_funcv0}
    Q_i'(v)=-\texttt{DistH}(v,g_i)-\lambda \texttt{Cost}(p_i(t),v)
\vspace{1.5mm}
\end{equation}
where $\lambda$ is a weight and $\texttt{Cost}(p_i(t),v)$ returns the action cost from state $p_i(t)$ to $v$.
We use the action cost similar to \cite{wen2022cl} where backward motions, turning, and changes of directions will receive additional penalty cost.
However, it's important to note that the distance heuristic is not a ``reachable" heuristic in the context of \clpibt. A heuristic is labeled as ``reachable" for $\pibt$-like algorithms if, while ignoring inter-robot collisions, a solitary robot is guaranteed reach its goal state by consistently choosing the ``best" action with the maximum $Q$ value at each timestep. In discrete \mpp, the single-agent shortest path length qualifies as a ``reachable" heuristic.
We mention that while Manhattan distance is a reliable ``reachable" heuristic when the map is obstacle-free, this is not guaranteed when obstacles are present. In scenarios with obstacles, a robot directed by the Manhattan distance heuristic might become entrapped in local minima, obstructing its path to the goal state.
Our case showed \texttt{DistH} is not a ``reachable" heuristic either. Despite being admissible, it lacks the required precision. Notably, the assertion that a greedy motion primitive should yield the highest $Q$ value among the possibilities isn't consistently upheld when using $Q_i(v)=-\texttt{DistH}(v,g_i)$.
This discrepancy easily leads the robot into a state of stagnation. 
Furthermore,  we found that this greedy heuristic could also lead to deadlocks.
An example can be seen in Fig.~\ref{fig:dead_lock}.

We present a novel count-based exploration heuristic to address the challenge, inspired by \cite{tang2017exploration}. This heuristic is specifically designed to surmount the issue of local minima arising from inaccuracies in distance heuristics and deadlocks and to foster exploration of uncharted territories.
To achieve this, we incorporate supplementary penalties for states that have been visited, aiming to encourage robots to break free from local minima and venture into unexplored regions. Each state $v=(x,y,\theta)$ is mapped to its nearest discrete state $\widetilde{v}=(\lfloor\frac{x}{\delta x}\rfloor,\lfloor\frac{y}{\delta y}\rfloor,\lfloor\frac{\theta}{\delta \theta}\rfloor)$. This mapping facilitates the tallying of occurrences using a hash table $\mathcal{H}_i$ for each robot, where $\delta x$, $\delta y$, and $\delta \theta$ delineate the state-space resolution.

When a robot $i$ traverses state $v$ at a given timestep $t$, the count $\mathcal{H}_i(\widetilde{v})$ is incremented by one. In addition, we introduce bonus rewards to incentivize the selection of the greedy motion primitive.
The resultant $Q$-function is:
\vspace{1mm}
\begin{equation}\label{eq: q_funcv1}
\begin{aligned}
Q_i(v)=Q'_i(v)+\alpha Gr(v)-\beta\mathcal{H}_i(\widetilde{v})
\end{aligned}
\vspace{1mm}
\end{equation}

Here, $\alpha$ and $\beta$ represent positive weight parameters. If the state $v$ results from a greedy motion primitive, $Gr(v)=1$; otherwise, $Gr(v)=0$.
To accommodate the requirement that robots should remain at their goal states upon arrival, we refrain from applying penalties to states near the goal states. As a result, when $v=g_i$ (the goal state of robot $i$), we set $\beta$ to zero.
We mention that such a $Q$ function could be ``reachable" if we properly choose the weight parameters.
Moreover, by implementing the count-based heuristic, which compels robots to explore distinct motion patterns for collision avoidance, the deadlock issues can be nicely achieved.
%
%

\section{Centralized \clecbs}\label{sec:ecbs}
We enhance the \clcbs algorithm by substituting the conventional low-level spatiotemporal hybrid state A* planner with focal hybrid state A* search \cite{barer2014suboptimal}. This refined approach is named \clecbs. In comparison to \clcbs, the \clecbs method guides the low-level planner to discover trajectories within a bounded suboptimality ratio while encountering fewer potential conflicts with other robots. 
This reduction in conflicts subsequently results in a significantly lower number of high-level expansions.

For the purpose of adapting \clecbs to lifelong scenarios, characterized by the necessity for frequent online replanning, we incorporate the windowed version of focal search in the low-level planning process \cite{li2021lifelong, han2022optimizing}. Specifically, during path planning, we only address conflicts that arise within a window of $\omega$ steps. This adjustment effectively decreases the runtime of the \clecbs algorithm during the planning phase. Replanning occurs at intervals of every $\omega$ steps and also when robots reach their current goals and receive new tasks.

\section{Evaluation}\label{sec:evaluation}
In this section,  we evaluate the proposed algorithms in static scenarios and lifelong scenarios.
All methods are implemented in C++. 
The source code can be found in \href{https://github.com/GreatenAnoymous/CarLikePlanning}{https://github.com/GreatenAnoymous/CarLikePlanning}.
%
All experiments are performed on an Intel\textsuperscript{\textregistered} Core\textsuperscript{TM} i7-6900K CPU at 3.0GHz in Ubuntu 18.04LTS.
In the simulation, the following parameters are used:
$w=2$, $l=3$, $l_b=2$, $\delta x=\delta y= 2$, $\delta\theta=40.1^{\circ}$, $u_{m}=2$, $\phi_{m}=40.1^{\circ}$, $r_m=3$, $\Delta t=r_m\delta \theta/u_{m}$.

\subsection{Static Scenarios}
In this section, we assess the performance of the algorithms on a grid of dimensions $100\times 100$, both with and without obstacles. For the scenarios involving obstacles, we create 50 circular obstacles with a radius of 1 unit length and place them randomly on the map.
For each value of $n$, we generate 50 distinct instances, ensuring that the states of the robot do not overlap with each other and with the obstacles in start and goal configurations. A time limit of 60 seconds is imposed on each instance.
The success rate is determined by tallying the instances each algorithm successfully solves within the time limit. Additionally, we evaluate the average runtime (in seconds), makespan, and flowtime across the solved instances.
Our evaluation includes a comparison with two reference algorithms: firstly, the prior centralized algorithm known as CL-CBS, and secondly, the decentralized SHA* algorithm as described in \cite{wen2022cl}. We emphasize that all algorithms use identical predefined motion primitives, except that GM is exclusively used for \clpibt.

A suboptimality ratio of 1.5 is selected for \clecbs.
Three variants of \clpibt are evaluated. The first, \clpibt(v0), employs the $Q$-function described in \eqref{eq: q_funcv1} without incorporating the count-based exploration heuristic.
Both \clpibt(v1) and \clpibt(v2) leverage the count-based exploration heuristic to resolve local minima and deadlocks.
In the case of \clpibt(v1), we opt to clear the hash table $\mathcal{H}_i$ and reset the associated counts to zero whenever robot $i$ reaches its designated goal state. Conversely, for \clpibt(v2), the visiting history is not cleared.
For all variants of \pibt, the maximum time step is constrained to 500.
And we set $\lambda=0.3$, $\alpha=\beta=\lambda \texttt{Cost}(p_i(t),v)$.

\begin{figure}[h]
    \centering
    \includegraphics[width=1\linewidth]{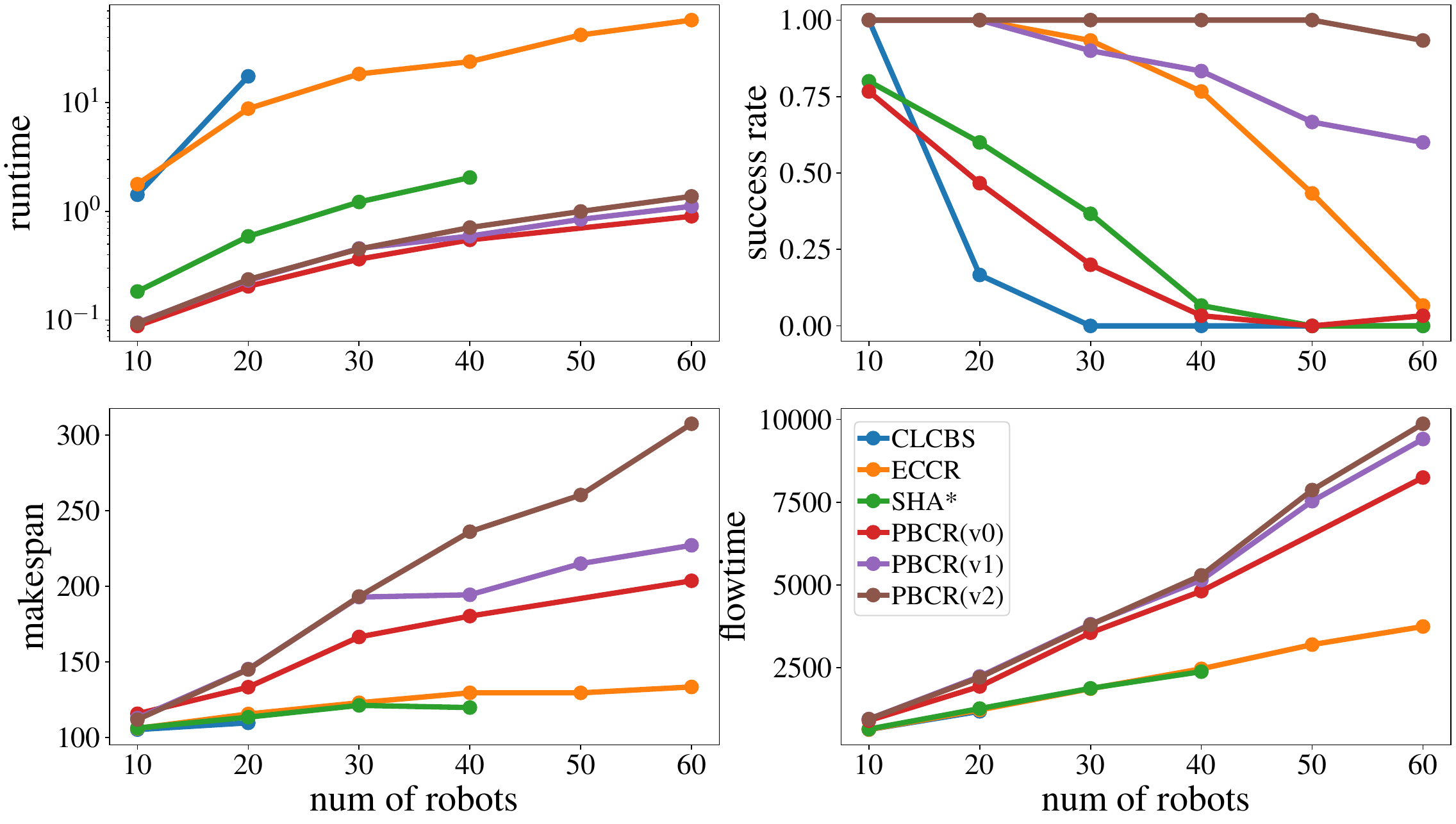}
\vspace{-5mm}
    \caption{Evaluation results on a $100\times 100$ empty map for a varying number of robots.}
    \label{fig:empty_oneshot}
\end{figure}

\begin{figure}[h]
\vspace{2mm}
    \centering
   \includegraphics[width=1\linewidth]{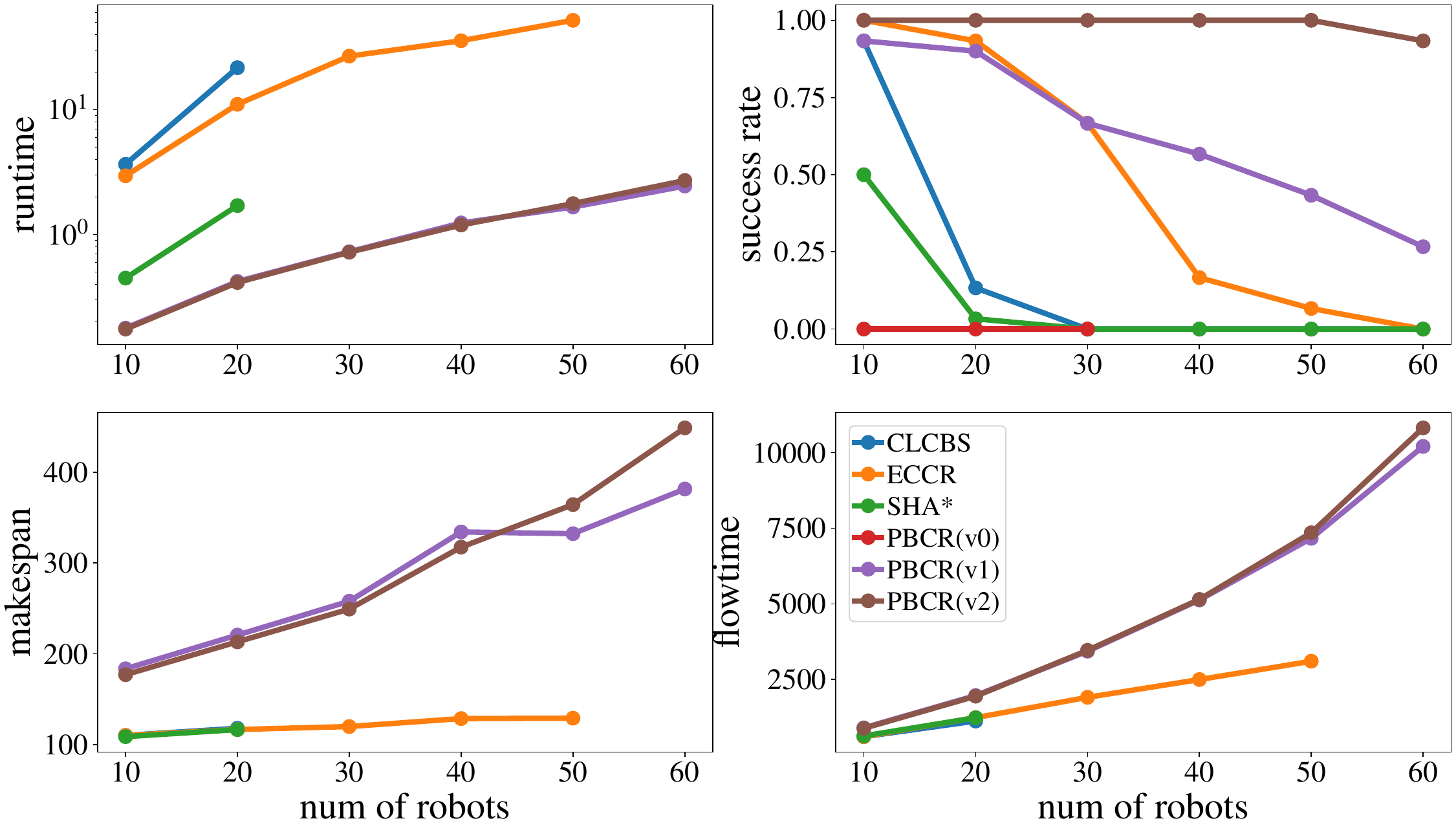}

\vspace{-3mm}
    \caption{Evaluation data on a $100\times 100$ map with $50$ randomly placed obstacles for a varying number of robots.}
    \label{fig:obstacle_oneshot}
\end{figure}

\begin{figure}[h]
    \centering
  \begin{overpic}               
        [width=1\linewidth]{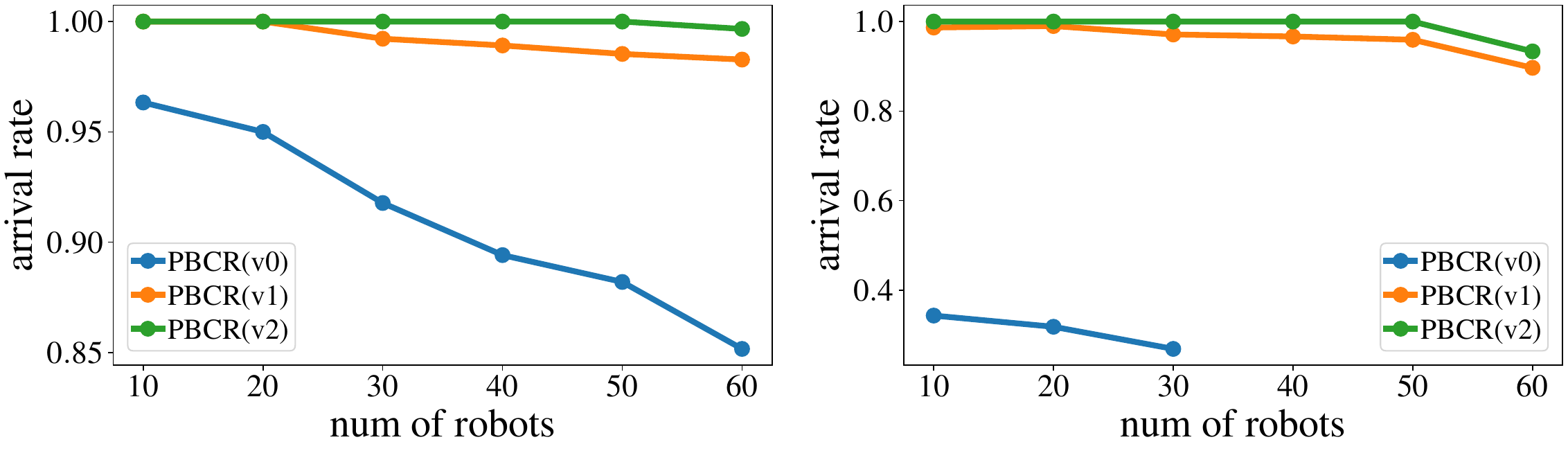}
             \put(25.5, -3) {(a)}
             \put(75.5, -3) {(b)}
        \end{overpic}
\vspace{-3mm}
    \caption{The average percentage of robots arrived at their goal state for each \clpibt variant on the $100\times 100$ maps for varying numbers of robots. (a) without obstacles (b) with obstacles.}
    \label{fig:arival_rate}
\end{figure}

Detailed evaluation results are presented in Fig.\ref{fig:empty_oneshot} through Fig.\ref{fig:arival_rate}.
Compared to CL-CBS, \clecbs showcases significantly enhanced scalability and success rates due to its capability to expand a notably smaller number of high-level nodes. Notably, the solution quality of \clecbs closely approaches that of CL-CBS.
On the flip side, the \clpibt variants excel in terms of runtime efficiency, resulting from their decentralized nature and one-step planning strategy. They can resolve instances involving 60 robots in as little as 4 seconds. Among these variants, \clpibt featuring the count-based exploration heuristic achieves an elevated success rate. Conversely, without the count-based exploration heuristic, \clpibt (v0) falters in instances with obstacles, revealing its limitations. However, \clpibt(v2) successfully tackles $93\%$ of cases involving 60 robots within 4 seconds, excelling in challenging scenarios—a substantial improvement over other variants and the previous decentralized SHA* approach.

It's crucial to emphasize that all \clpibt variants do not fail due to time constraints but rather due to surpassing the maximum timestep.
For those failed instances, \clpibt with count-based heuristic still can guide more than $90\%$ of the robots to their goal states as shown in Fig.~\ref{fig:arival_rate}.
The contrast in success rates between \clpibt(v2) and \clpibt(v1) implies that retaining the visiting history until completion, instead of resetting it upon each robot's arrival at its goal state, is more effective.
This arises from the fact that in the context of \clpibt, when a robot reaches its designated goal, it often has to temporarily vacate its goal position to accommodate other robots that must traverse it. This frequent shifting in and out of the goal state can potentially lead to repetitive cycles and hinder progress.
Maintaining the visiting history mitigates this issue by preventing the robot from becoming trapped in an endless loop of entering and leaving the goal, consequently boosting the overall success rate.
One drawback of \clpibt lies in its trajectory quality. This stems from the absence of global information and the inherent nature of one-step planning, leading to longer planned trajectories compared to centralized algorithms such as \clecbs and CL-CBS. This effect is particularly pronounced when dealing with a large number of robots.

\subsection{Lifelong Scenarios}
\begin{figure}[h]
    \centering
  \begin{overpic}               
        [width=1\linewidth]{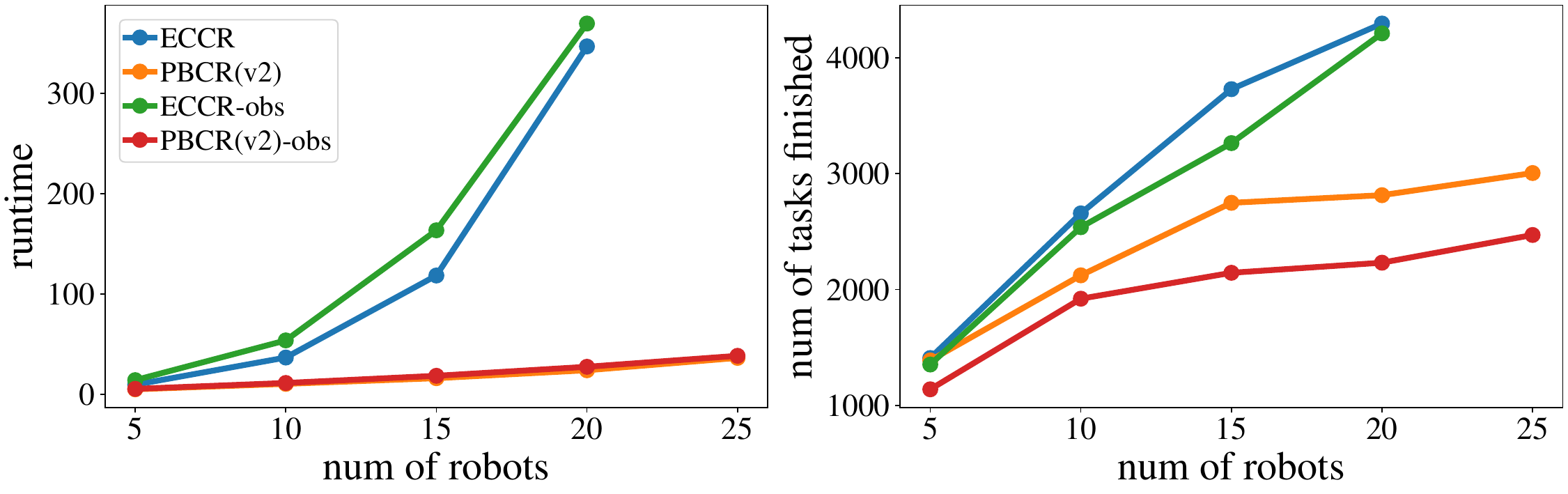}
        \end{overpic}
\vspace{-5mm}
    \caption{Runtime and the average number of tasks finished within given timesteps in lifelong scenarios for varying number of robots. The term ``obs'' is an abbreviation for scenarios with obstacles.}
    \label{fig:lifelong}
\end{figure}

In this section, we subject both \clecbs and \clpibt to testing within lifelong settings.
We adopt a $50 \times 50$ map configuration, both with and without obstacles. For scenarios involving obstacles, a set of 10 obstacles is distributed randomly.
For each value of $n$, we randomly generate 20 unique instances. In each instance, both the initial configurations and 4000 goal states, randomly generated, are allocated to each robot.
Throughout the simulation, we assume a lack of a priori knowledge on the part of the robots regarding their subsequent tasks.
In each instance, we define a maximum of 5000 simulated steps and set a runtime limit of 600 seconds. Each robot has many tasks that cannot be feasibly completed within the designated maximum steps.
Regarding \clecbs, a suboptimality ratio of 1.5 is established, while a window size $\omega$ of 5 is employed.
Given that new goals are allocated to robots upon completing current tasks, the visiting history is consistently cleared. This decision is driven by the fact that the visiting experience only relates to the robot's preceding task.
\clpibt(v0) is excluded from consideration owing to its poor performance in addressing issues related to deadlocks and stagnation.
The outcome of our evaluations is depicted in Fig.~\ref{fig:lifelong}.
\clpibt achieves significantly lower execution times than \clecbs, rendering it suitable for handling large-scale scenarios and real-time planning. Nonetheless, it accomplishes a smaller number of tasks.

\subsection{Real-Robot Experiments}
\begin{figure}[h]
\vspace{2mm}
    \centering
  \begin{overpic}               
        [width=\linewidth]{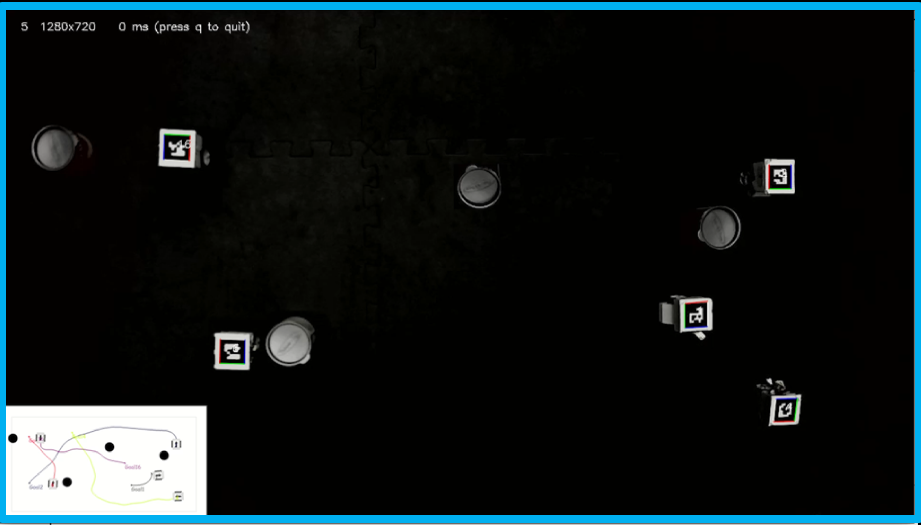}
        \end{overpic}
    \caption{Snapshot of a real robot experiment with 4 obstacles and 5 robots.}
    \label{fig:micromvp}
    \vspace{-2mm}
\end{figure}

We employed the portable multi-robot platform, microMVP \cite{yu2017portable}, for conducting real-world experiments involving car-like robots to validate the execution of our algorithmic solutions on real hardware. Whereas the microMVP robots are differential drive robots, a software layer can be imposed to simulate car-like robots, which is what we did \footnote{While achieving the same goal, the simulation makes the experiment more challenging than running directly on car-like robots.}. 
Fig.~\ref{fig:micromvp} gives a visual representation of the setup. As evident from the attached video, paths generated by our planners can be successfully executed on the robots' controllers.

\vspace{-1mm}
\section{Conclusion and Discussions}\label{sec:conclusion}
In this study, we investigate path planning for multiple car-like robots. We present two distinct algorithms tailored to the specific demands of car-like robot navigation scenarios.
The first and our main contribution, \clpibt, employs an effective count-based exploration heuristic. Focusing on decentralized decision-making, \clpibt demonstrates a notable advancement over previous decentralized approaches. The improvement is evident through significantly heightened success rates with some manageable optimality trade-offs of yielding longer trajectories. 
Opportunities for improving trajectory quality within \clpibt remain; in particular, our approach employs a manually designed $Q$-function for action selection, which may be replaced with data-driven approaches for reaching optimal performance.

\bibliographystyle{formatting/IEEEtran}
\bibliography{bib/all}

\end{document}